\documentclass[runningheads]{llncs}
\usepackage{hyperref}
\usepackage{url}
\usepackage{graphicx}
\usepackage{doi}
\usepackage{graphicx}
\usepackage{xcolor}
\usepackage{algorithm}
\usepackage{algpseudocode}
\usepackage{tikz,amsmath} 
\usepackage{subcaption}
\usepackage[normalem]{ulem}
\begin{document}
\title{Vertical Federated Image Segmentation}
%
%
\author{Paul K. Mandal\inst{1}\orcidID{0000-0002-4966-0494} \and
Cole Leo\inst{2}\orcidID{0009-0008-6875-6281}}
\authorrunning{P.K. Mandal and C. Leo}
%
\institute{Department of Computer Science, The University of Texas at Austin, Austin TX 78712, USA \email{mandal@utexas.edu} \and
T2S-Solutions, Aberdeen MD 21017
\email{cleo@t2s-solutions.com}}
\maketitle              
\begin{abstract}
    There has been a growing concern for both data privacy and acquisition for the purpose of training robust computer vision algorithms. Often, information is located on separate data silos and it can be difficult for a machine learning engineer to consolidate all of it in a fashion that is appropriate for model development. Additionally, some of these localized data regions may not have access to a labelled ground truth, rendering conventional model training impossible. In this paper, we propose a novel architecture that can perform image segmentation in vertical federated environments. This is the first implementation of a federated architecture that can perform vertical federated image segmentation. We utilized a distributed vertical fully convolutional network architecture that is able to train on data where the segmentation maps reside on a different federate than the original image. Our architecture is able to compress the features of an image from 49,152 down to 500, allowing for more efficient communication between the top and bottom model of our federated architecture. We trained our model on 369 images from the CamVid dataset. Our model demonstrates a robust capability for accurate road detection.

\end{abstract}
\section{Introduction}
\label{sec:intro}

The growing popularity of machine learning in a diverse set of domains requires substantial amounts of data to train these algorithms. 
Thus, there has been a growing interest in procuring this information . However, a common issue is that the data required to solve these complex problems is often scattered amongst a variety of locations,
resulting in the inability to perform model development locally as standard modelling practices are unable to utilize dispersed information. Even if the desired contents for training were sent via a network schema, there would still be hefty delays in the development process and would warrant issues if problems with connectivity occurred. This also does not account for potential security breaches that could arise from a faulty network, and as a result distributed learning solutions have been exploited to combat scenarios such as these.  \cite{9562559,10.1145/3377454}.  

Federated Learning (FL) is a distributed machine learning approach where a model is trained across decentralized devices or servers holding local data samples without directly exchanging them \cite{10.1145/3450288,Liu_2022}. The model is trained collaboratively, aggregating insights from various sources while preserving data privacy and security.
There exist three main categories within the domain of federated learning, these being horizontal (HFL), vertical (VFL), and transfer (FTL) paradigms, wherein each maintains the same theory of distribution but differ in their exact capabilities \cite{10.1145/3298981}. The variation with the most robust research backing is HFL, that of which operates on parties that share the same feature space but contain differing samples \cite{9874186}. An example of this would come in the form of a banking or insurance firm that record a set of statistics about a given consumer, that has since spread from one location to another and now finds itself analyzing new users \cite{Long2020}.

However, our paper is focused on VFL which has recieved less reasearch attention than HFL \cite{feng2020multiparticipant}. For HFL, information held among the parties differs in the feature space rather than the sample entries, indicating that most of the participating federates do not have access to the labels required to carry out training \cite{liu2023vertical,Wu_2020}.
VFL allows multiple parties with different feature spaces are able to train a shared, distributed ML model where each party holds a different part of the model \cite{mammen2021federated}. In this paper, we present a novel vertical federated FCN that is distributed across two different federates, where the first federate contains the images and the second federate contains the segmentation maps. We provide a literature review in section \ref{sec:Background}, outline our experimental setup and model architecture in section \ref{sec:Methodology}, and discuss our results in section \ref{sec:Results}. We finally enumerate future work in section \ref{sec:Discussion}.
\section{Background}
\label{sec:Background}

\subsection{Image Segmentation \& Fully Convolutional Networks}

Image segmentation is a type of computer vision problem where each pixel of an image is assigned a class that it belongs to \cite{minaee2020image}. One of the most effective ways to operate on this information is with Fully Convolutional Networks (FCNs). FCN's have historically yielded the highest performance on image segmentation problems \cite{long2015fully}. FCN's follow an encoder-decoder like structure where convolutions are used to decrease the number of features followed by convolutional transpose layers to expand the dimensionality.

There have been state of the art results with attention in image segmentation \cite{xie2023attention}. Additionally, self guided attention mechanisms for image segmentation look extremely promising \cite{sinha2020multiscale}. However, we have opted to use FCN's in this paper for two main reasons: (1) We wanted to keep the model simple and (2) we wanted to be able to use VGGNet as our model base.

\subsection{Vertical Federated Learning}
Most of the literature on Federated learning pertains to Horizontal Federated Learning (HFL) \cite{9874186}. This is where each federate has a shared feature space but different training examples. In HFL, models can be trained locally, and then the arbiter simply takes a weighted average of the model weights \cite{DBLP:journals/corr/McMahanMRA16}.

\begin{figure*}
    \centering
    \includegraphics[scale=2,width=0.9\textwidth]{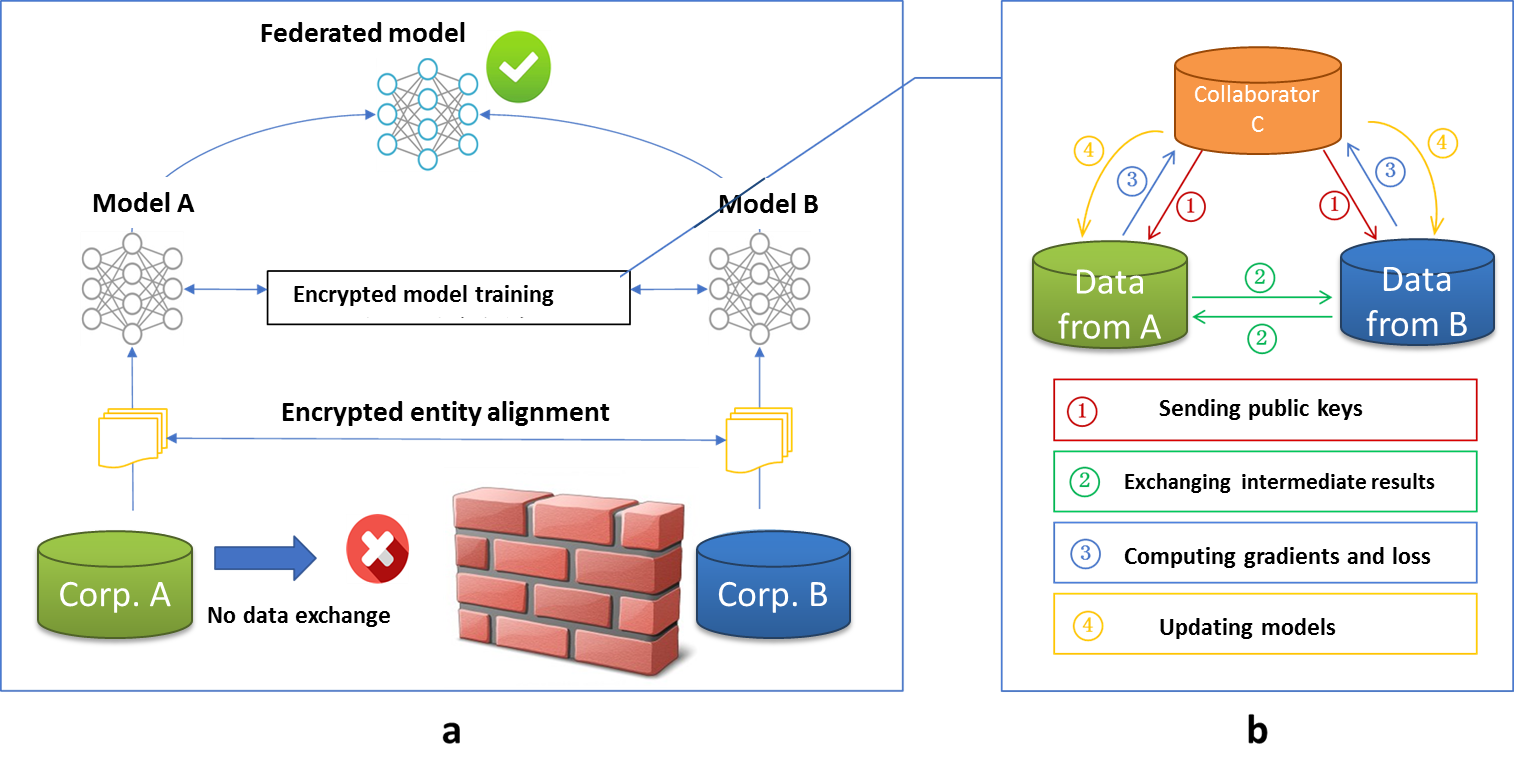}
    \caption{Vertical Federated Learning Architecture as proposed in \cite{10.1145/3298981}}
    \label{fig:vfl_arc}
\end{figure*}

The emphasis of our paper is on Vertical Federated Learning (VFL), wherein each federate encompasses a distinct feature space \cite{10.1145/3514221.3526127,liu2022vertical}. The training process of VFL is much less intuitive than HFL. In a VFL problem, a federates local inputs are propagated through a bottom model. The outputs of the bottom model are encrypted and then sent to an interactive layer. The interactive layer performs entity alignment for each training example across all of the different federates and the label. From here, these compressed features can now be forwarded to the top model. At this point, only the top model sees the label, and backpropagation occurs as it would for any other model. When these losses reach the interactive layer, they are encrypted and sent to their respective federate's bottom models \cite{9005992}. The architecture of a typical vertical federated learning scenario is shown in figure \ref{fig:vfl_arc}.

Although VFL is more complicated than HFL, VFL has several interesting properties. Since different federates do not share the same featurespace, bottom models do not need to be identical across different federates. Thus, if Federate A contains image data and Federate B contains text data, it is entirely possible to create a multimodal model where Federate A's bottom model is a CNN and Federate B's model is an attention model or LSTM \cite{mdpiMFL}.


\section{Experimental Setup}
\label{sec:Methodology}

Our experiment was structured in the way that a conventional VFL model would be ran. Our images and bottom models resided on one federate, and our intermediate layer, top layer, and labels (image segmentations) ran on another.

\subsection{CamVid Dataset}
CamVid, short for Cambridge-driving Labeled Video Database, is a popular dataset used in computer vision and machine learning research \cite{BROSTOW200988}.
It consists of images captured from a vehicle driving through urban and suburban environments. The dataset is primarily used for semantic segmentation tasks, where the goal is to classify each pixel in an image or frame of a video into a specific object class, such as road, car, pedestrian, or building. CamVid is valuable for developing and testing algorithms related to autonomous driving, scene understanding, and image analysis.

For our experiments, we only did a binary classification to detect roads. The purpose for this was to reduce the dimensionality of data that was required to be sent between federates, as this is the most time consuming part of VFL.

\subsection{Bottom Model for Convolutions}
\label{sec:bottom}
The first item in the network architecture is the bottom model, or the stage wherein convolutions are applied to reduce dimensionality of the input sample. This is the most common component, and will be seen on the majority of the data silos present in the system. As it only operates on data within the determined sample space, there is no need for labelled features to be present and is thus capable of operating on the available information. 

\begin{figure}[H]
    \centering
    \includegraphics[scale=2,width=0.8\textwidth]{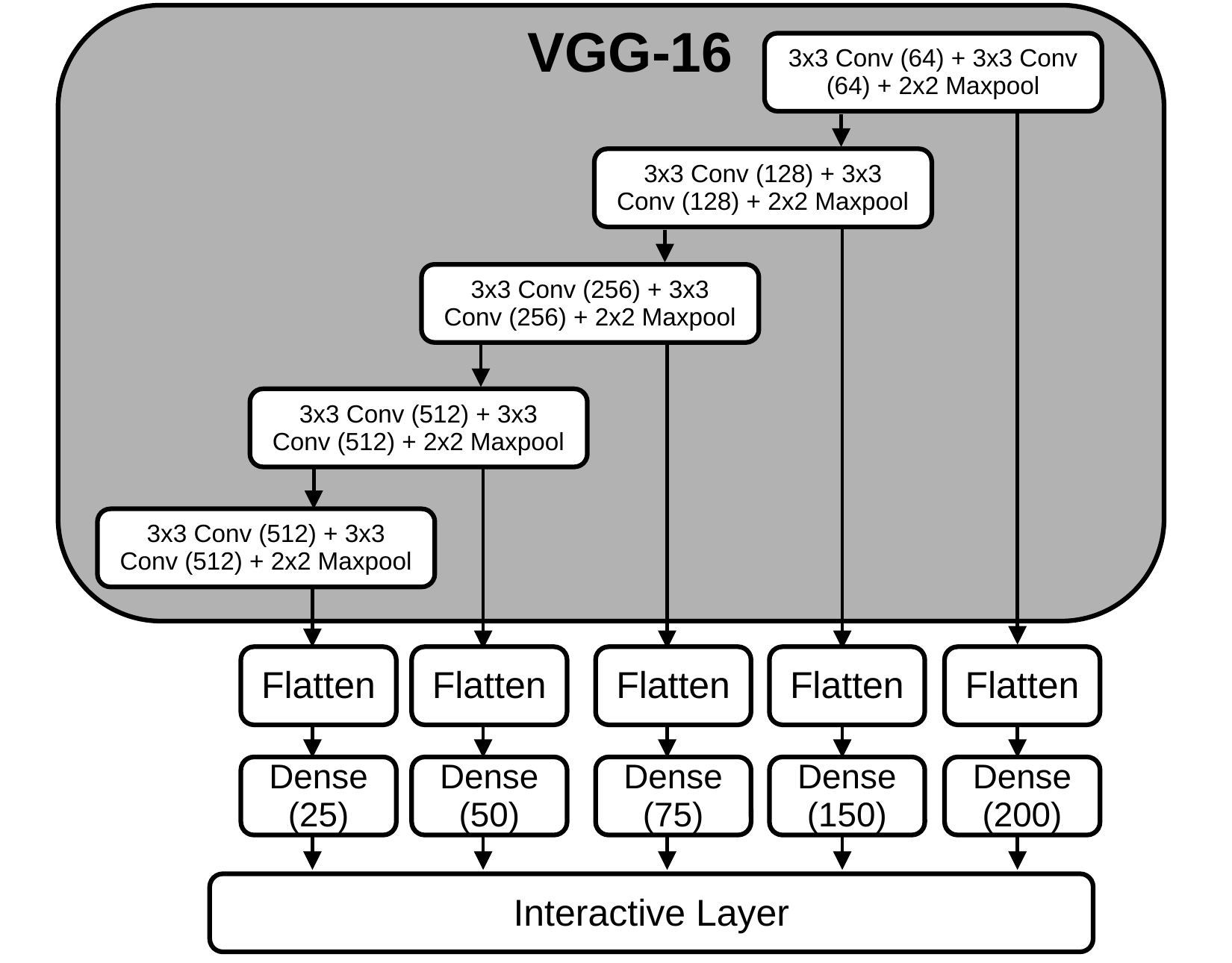}
    \caption{Architecture of our bottom model.}
    \label{fig:bottom}
\end{figure}

The VGG16 pretrained model was used as the core structure for the backbone of our FCN \cite{wang2015places205vggnet}. By applying a sequence of convolutions followed by a set of normalization and pooling paradigms, it is possible to drastically compress the original sample and prepare it for the interactive layer. Perhaps the most important component out of these operations is the max pooling layers applied after every step, as their outputs are saved separately in order to be used as a pretrained weight in the top model. By doing so, higher accuracies can be achieved within a sliced network as a result of maintaining pertinent information about the training process occurring on the individual federates. This can also help prevent over and underfitting that may result from operating on sparse data. The architecture of our bottom model is shown in figure \ref{fig:bottom}.

\subsection{Dimensionality in the Interactive Layer}
\label{sec:interactive}
A critical component in a VFL model comes in the form of the interactive layer. It serves as the point in the network architecture wherein the aggregated features of the individual federates are served to the central server for classification. Here, the data must be augmented in such a way that preserves information all the while undergoing drastic dimensionality compression. When working with standard RGB imagery, there can often be upwards of over a million features that must be processed and sent over the shared network. Thus, we had to downscale our images down to 128x128 pixels.

In an uncompressed scenario that yields 1000 features, the time to train per epoch is roughly four hours due to network constraints. This grows exponentially as the number of features increases. Thus, it becomes necessary for the interactive layer to consolidate these features in order to adhere to the observed limitations. Our architecture manages to compress the number of features in our images from 49,152 to 500, which is about 1\% of the original image size.

Hinton and Salakhutdinov first proposed using linear layers instead of PCA for dimensionality reduction \cite{hintonDimension}. Performing this level of reduction cannot be achieved by simply applying a sequence of flatten and linear layers that provide the desired feature count, as doing so does not account for the weights present in the skip convolutions of the bottom model. An innovative means of preserving these parameters is to apply the aforementioned sequence on each of the individual skip convolution outputs rather than the result of the model. Being that each of these operations results in a different number of features, it is necessary to scale the number of items being compressed on a per convolution basis. With this being the case, each of the five major cross-correlations were filtered and represented as 25, 50, 75, 150, and 200 features respectively. By doing so, it is possible to maintain pertinent information about the weights for each operation being conducted on the bottom model with respect to the dimensionality of their outputs. 

\subsection{Top Model for Classification}
The final portion of the vertical FCN can be recognized as the classification model present on the central serving medium. Here, label information that is not available on the individual federates operating within the system is located, prompting the model training process to finalize and reach deterministic conclusions. This intuitive network accounts for the transgressions of the previous interactive layer, augmenting the outputs of the linear layers in such a way that they can be rebuilt and classified through a series of convolutional transpose and cross-reference layers.

It is first necessary to present the heavily compressed weights to the transpose network in a way that accounts for image dimensions, number of channels, and the desired batch size. Being that these values were consolidated to the aforementioned sizes in section \ref{sec:interactive}, they must be expanded upon in order to adhere to their original dimensionalities in the bottom model. It is possible to obtain the necessary tensor shapes by applying linear filters to the appropriate compressed weights and subsequently passing them through an unflatten schema. These layers are modified in a way that reflects the original pretrained weights and can then be cross-referenced by the transpose network.

Upon performing all of the necessary reshaping of information obtained by the interactive layer, the data can be passed through the remainder of the FCN architecture. This consists of a sequence of transpose convolutions, or deconvolutions that operate on the information to increase the dimensionality of the input \cite{jimaging7100210}. In doing so, the output tensor begins to mimic the sizing conventions of the labelled ground truth present on the federate. Amid the calculations for each of these cross-references, the uncompressed weights are applied to the output, thus factoring in the results of the previous skip convolutions. After accounting for each of the shared parameters, a final convolution is swept across the resultant image in order to determine the class of each image segment with respect to the label data.

\begin{figure}
    \centering
    \includegraphics[scale=2,width=0.9\textwidth]{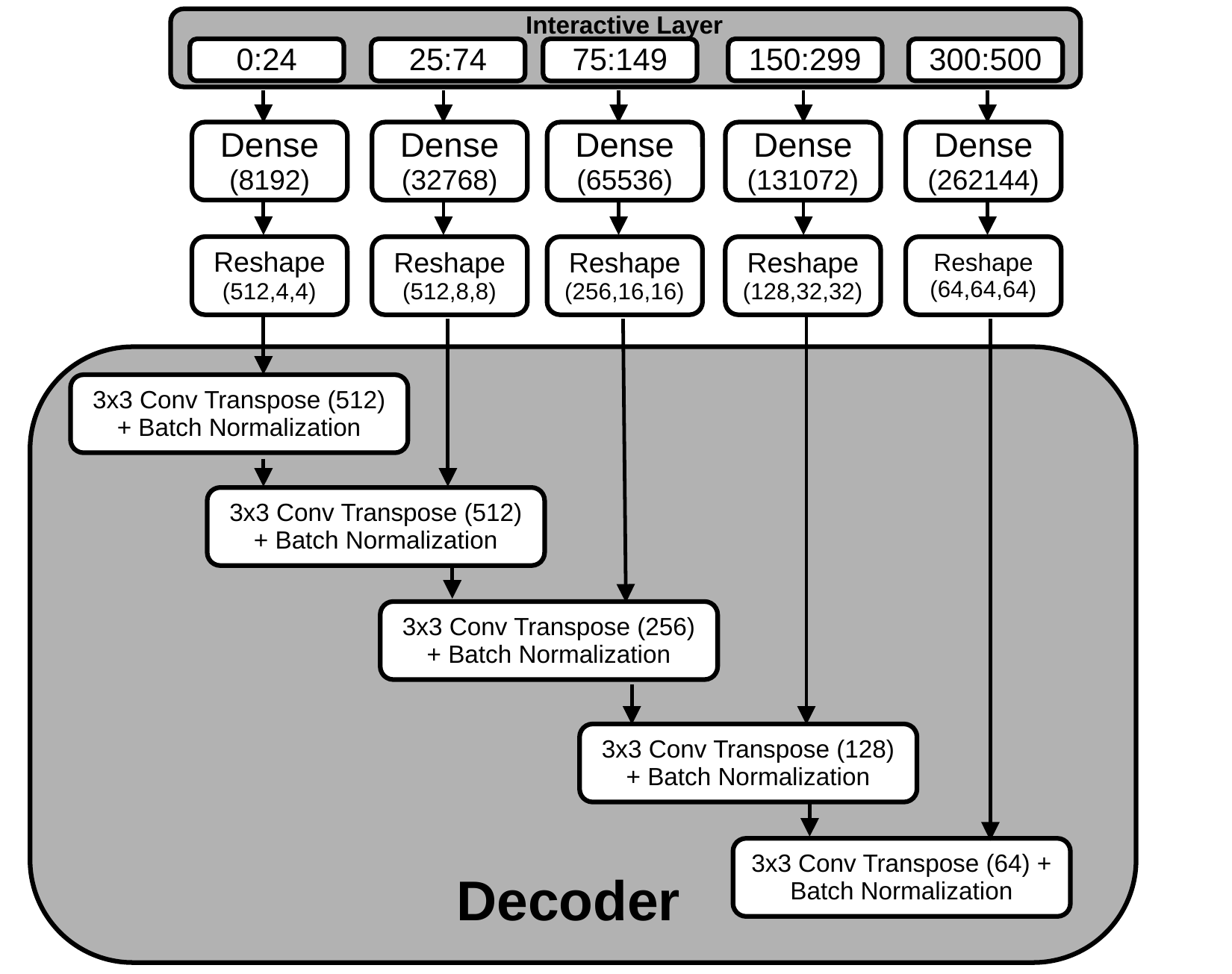}
    \caption{Architecture of our top model.}
    \label{fig:top}
\end{figure}

\subsection{Modifications to FATE}
\label{sec:fatemods}
In the course of our research, we encountered a limitation in the FATE architecture that posed a challenge to our specific use case. FATE is not designed to handle use cases where Federate A exclusively contained sample data, and Federate B exclusively held labeled data. This presented a significant hurdle, as the FATE approach presupposes that the federate containing labeled data also maintains sample data, which was not the case in our scenario.

To solve this, we implemented a mechanism for accessing the encrypted sample data in Federate A and retrieving the decrypted values from the interactive layer. This allowed us to facilitate the training process on appropriate data in FedB, thereby circumventing the challenge of operating on 0 tensors.

Furthermore, when assessing performance using the FATE approach, a critical drawback emerged – the unavailability of intuitive metrics due to the restricted access to 0 tensors. In response, we implemented a solution to conduct validation metrics during runtime, leveraging the decrypted data from the interactive layer as the reference point.

To streamline this, we created a system capable of running metrics such as Intersection over Union (IoU) via Jaccard Index and accuracy (Acc) using a custom class. These metrics were computed on batches from each training iteration, and the respective values, along with batch contents, were saved for post-process validation. This not only provided a more comprehensive understanding of model performance but also ensured that the lack of access to certain tensors in the FATE architecture did not impede our ability to derive meaningful insights during the evaluation phase.
\section{Results}
\label{sec:Results}

\begin{figure}
    \centering
    \includegraphics[scale=2,width=0.45\textwidth]{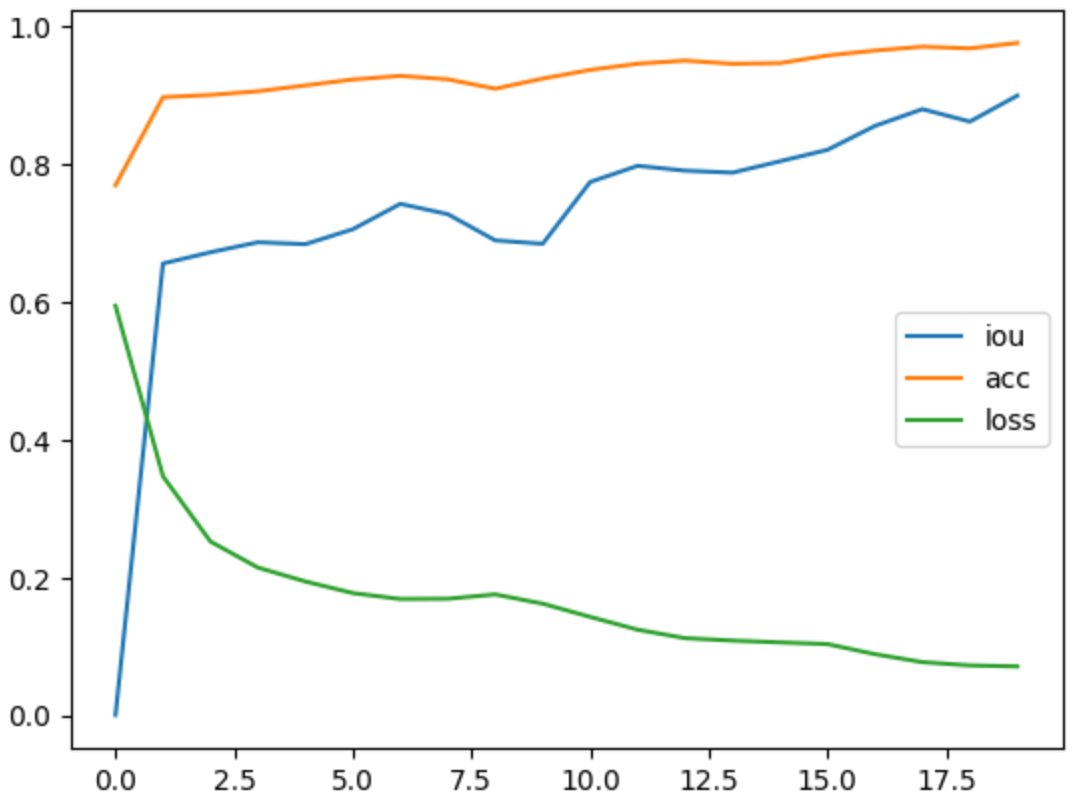}
    \caption{Pixel Accuracy, IOU, and loss on the training set.}
    \label{fig:graph}
\end{figure}

The performance metrics on our training set demonstrated promising results. On our training set, our model achieved a 95\% pixel accuracy, while the Intersection over Union (IOU) metric hovered around 90\%. It is crucial to note that these evaluations were for binary classification, distinguishing it from the multiclass scenario of the CamVid dataset. The performance is shown in figure \ref{fig:graph}.

\begin{figure}
  \centering
  \includegraphics[height=1in]{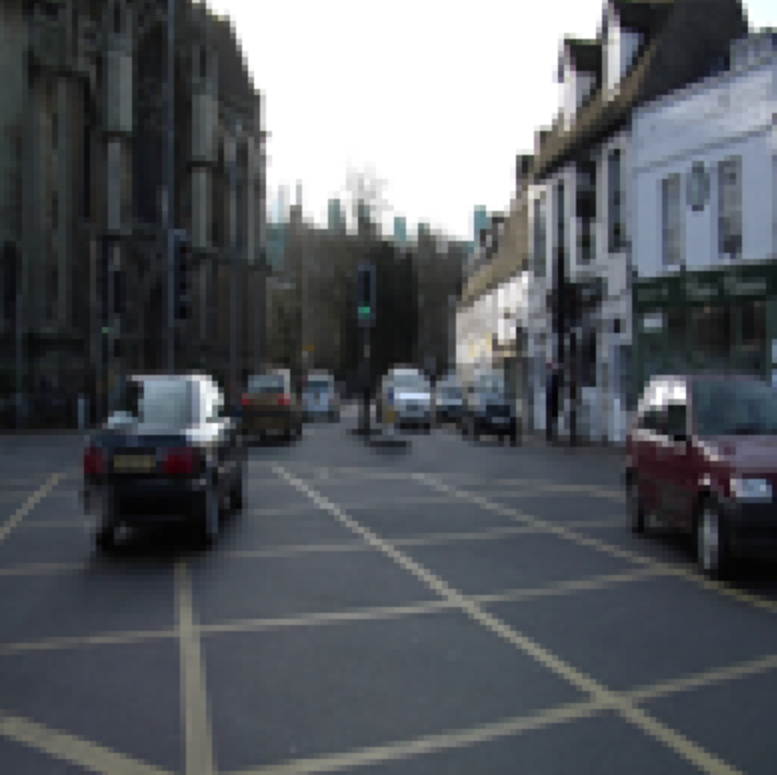}\space
    \includegraphics[height=1in]{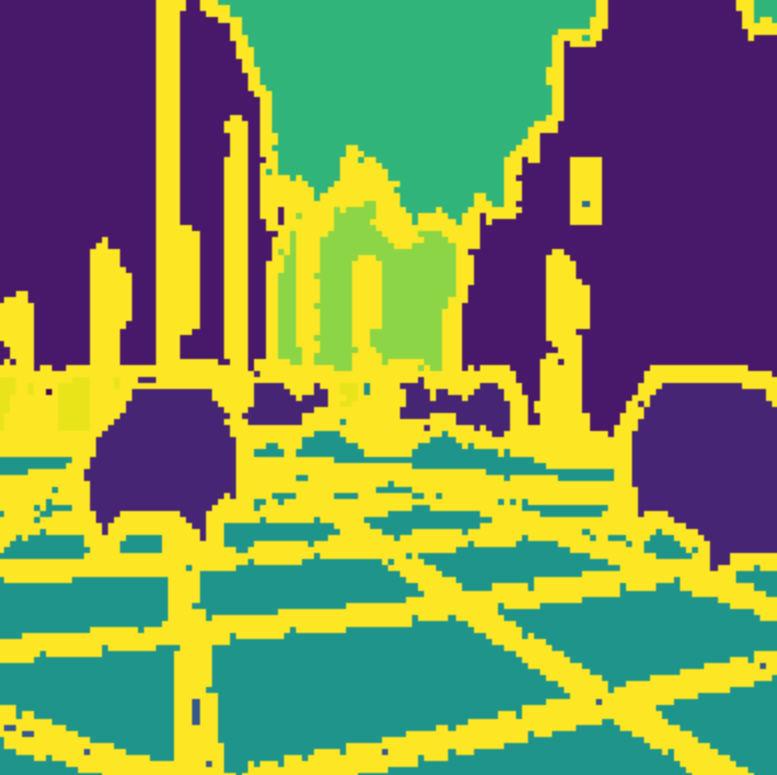}\space
    \includegraphics[height=1in]{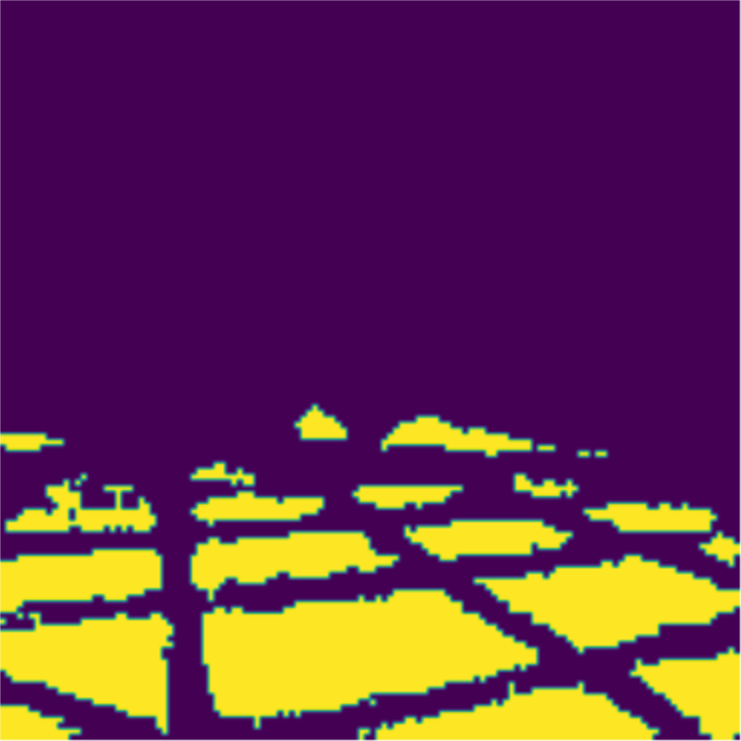}
  
  \includegraphics[height=1in]{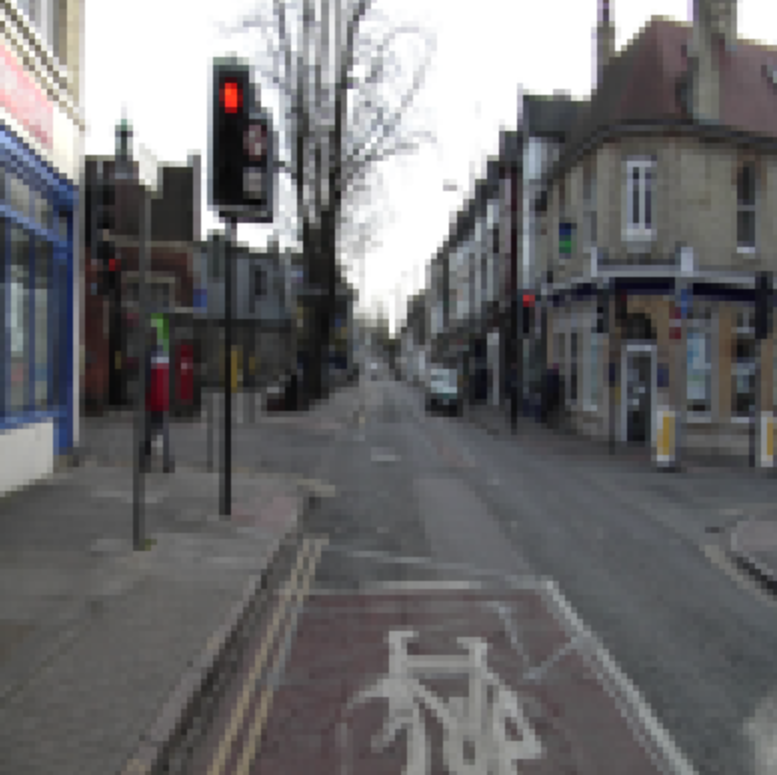}\space
    \includegraphics[height=1in]{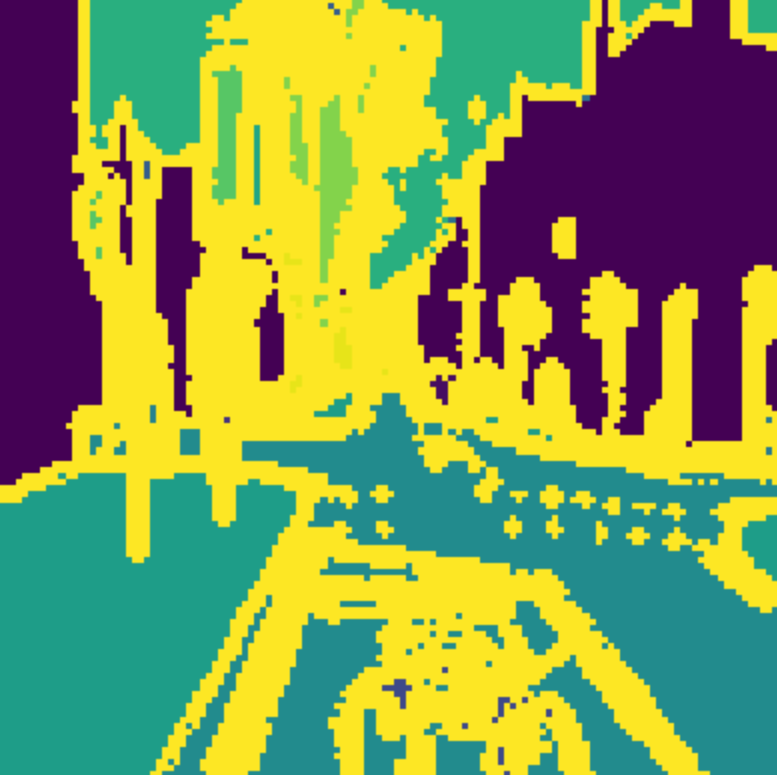}\space
    \includegraphics[height=1in]{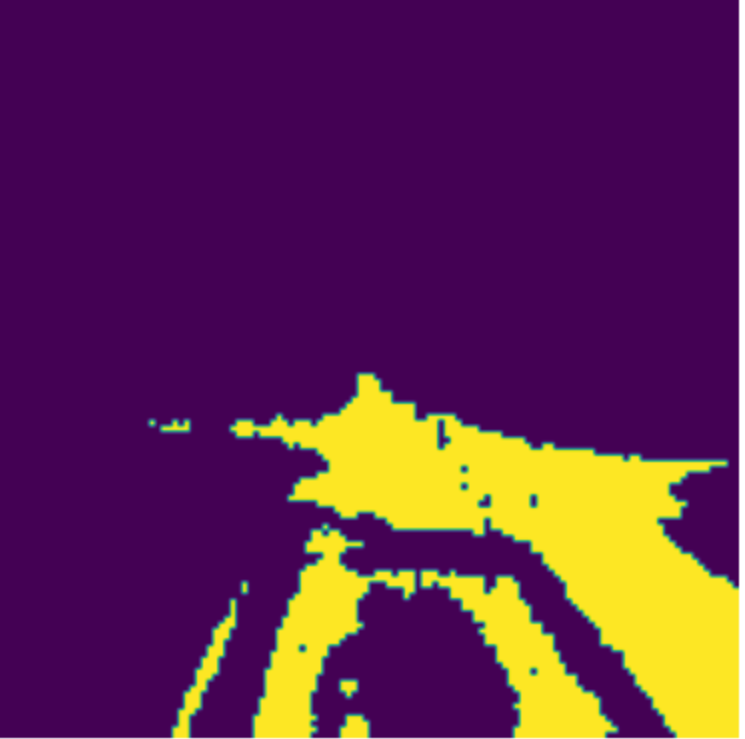}

  \includegraphics[height=1in]{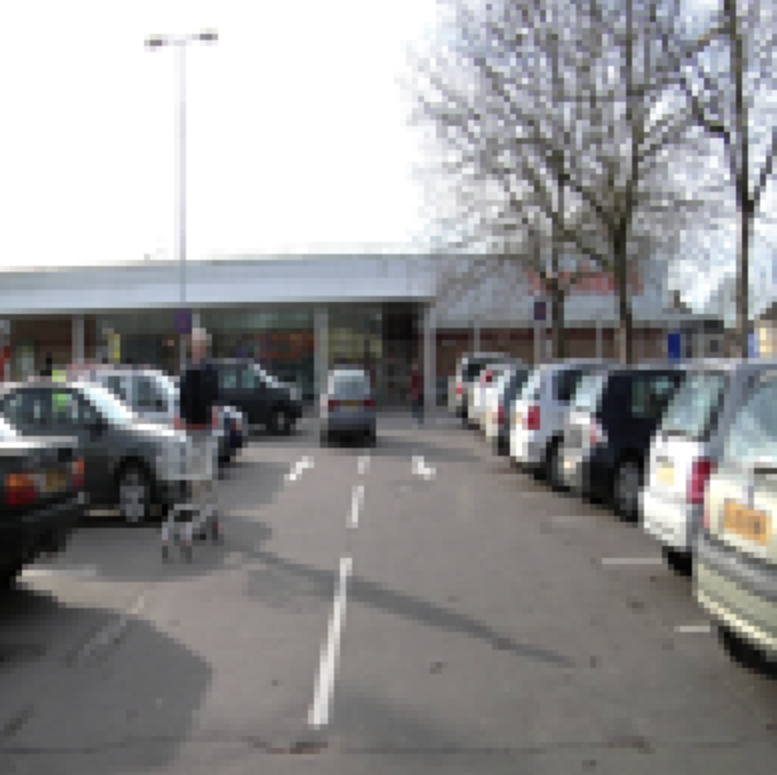}\space
    \includegraphics[height=1in]{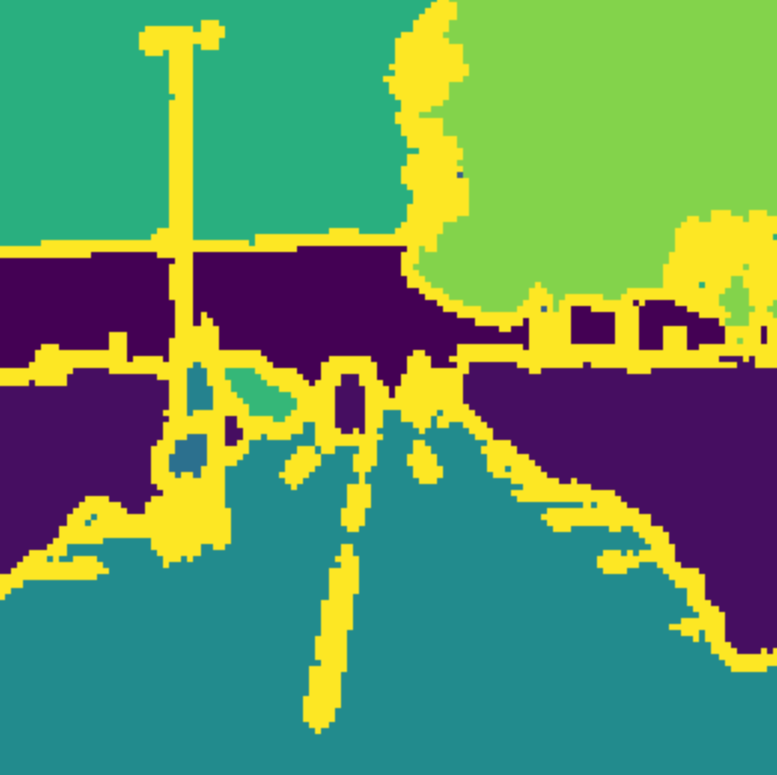}\space
    \includegraphics[height=1in]{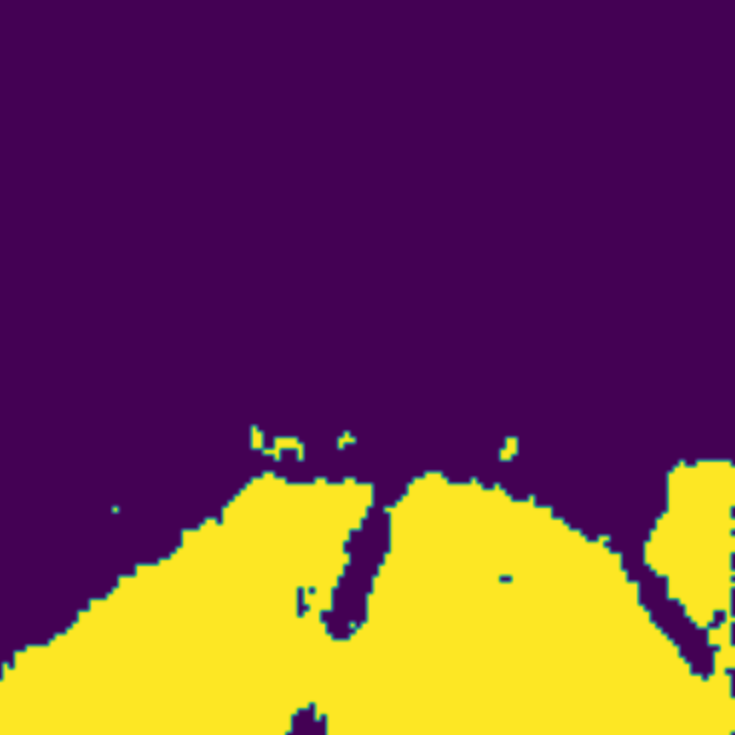}
    
  \caption{Left - The original image; Center - The segmentation map; Right - Our model's output}
  \label{fig:results}
\end{figure}

Unfortunately, FATE does not support supplying computation of validation set during training. Consequently, we were unable to generate plots illustrating the trajectory of validation loss, pixel accuracy, and IOU over time. Despite this limitation, our model's capabilities are exemplified through example outputs from the test set, as depicted in figure \ref{fig:results}. These visual representations showcase the model's proficiency in producing sensible segmentation results, providing valuable insights into its practical application.

\section{Discussion \& Future Work}
\label{sec:Discussion}

Although the method for feature compression employed in our model for vertical federated image segmentation yields satisfactory results, there is likely significant room for improvement. Improving feature compression is critical for achieving faster training times and enhancing the robustness of the model. This optimization is particularly important for expanding the model's capabilities to handle multiclass classification and process larger images.

Future work should focus on exploring and refining feature compression techniques to achieve superior efficiency in our model. This optimization aims to not only accelerate training times but also improve the overall resilience of the model. The success of this optimization is foundational for extending our model's applicability to more complex scenarios and larger datasets such as Cityscapes \cite{cordts2016cityscapes}. We attempted to write a vertical federated Faster Region-based Convolutional Neural Network. However, we encountered issues with handling how the region proposals are sent through the interactive layer. We believe that a vision transformer would allow us to circumvent this issue, and we intend to explore this in a future paper.

Furthermore, the expedited training times resulting from enhanced feature compression present the opportunity to work with larger datasets. Access to larger datasets contributes to a more comprehensive understanding of data patterns, enhancing the model's generalization capabilities. The combination of improved feature compression, faster training, and larger datasets is expected to significantly enhance the overall performance and adaptability of our model.

In summary, while the current model demonstrates commendable performance, the focus of future research should be on technical refinements in feature compression. This optimization is crucial for achieving faster training, increasing model robustness, and enabling the model to handle more complex tasks and larger datasets in vertical federated image segmentation.
%
%
%
\bibliographystyle{splncs04}
\bibliography{mybibliography}
\end{document}